# A Sentence is Worth a Thousand Pictures: Can Large Language Models Understand Hum4n L4ngu4ge and the W0rld behind W0rds?

Evelina Leivada[1,2], Gary Marcus[3], Fritz Günther[4], Elliot Murphy[5]

1. Universitat Autònoma de Barcelona, Barcelona, Spain
2. Institució Catalana de Recerca i Estudis Avançats (ICREA), Spain
3. New York University, New York, USA
4. Humboldt-Universität zu Berlin, Berlin, Germany
5. Vivian L. Smith Department of Neurosurgery, University of Texas Health Science Center at Houston, Texas, USA

**Abstract:** Modern Artificial Intelligence applications show great potential for language-related tasks that rely on next-word prediction. The current generation of Large Language Models (LLMs) have been linked to claims about human-like linguistic performance and their applications are hailed both as a step towards artificial general intelligence and as a major advance in understanding the cognitive, and even neural basis of human language. To assess these claims, first we analyze the contribution of LLMs as theoretically informative representations of a target cognitive system vs. atheoretical mechanistic tools. Second, we evaluate the models' ability to see the bigger picture, through top-down feedback from higher levels of processing, which requires grounding in previous expectations and past world experience. We hypothesize that since models lack grounded cognition, they cannot take advantage of these features and instead solely rely on fixed associations between represented words and word vectors. To assess this, we designed and ran a novel 'leet task' (l33t t4sk), which requires decoding sentences in which letters are systematically replaced by numbers. The results suggest that humans excel in this task whereas models struggle, confirming our hypothesis. We interpret the results by identifying the key abilities that are still missing from the current state of development of these models, which require solutions that go beyond increased system scaling.

## Introduction

Recent advances in Artificial Intelligence (AI) have given rise to many applications that have captured the public imagination (e.g., ChatGPT, DALL·E 2/3, Imagen, Stable Diffusion). These applications can create realistic images and/or synthetic language, based on textual prompts, while others are fluent text generators. This dual ability to process natural language and produce synthetic language that looks remarkably like that of humans, has given rise to claims that AI has passed the *language threshold*. Some AI applications have been linked to claims of passing the Turing test (Oremus 2022), of showing human-like understanding and sentience (Lemoine 2022), and of being able to extrapolate the morphophonological rules of natural language in a way that can be compared with language learning by children (Ellis et al. 2022).

Yet, good tools are not synonymous with good models that provide faithful representations of the target system. In general terms, models aim to capture some aspects of a target system with the aim to help construct a theory of it. For this to happen, the models need to be reliable and transparent representations of the target system. Given their stochastic nature (Bender et al. 2021), at present it is not yet clear what LLMs can tell us about human language. Indeed, some researchers prefer the term ‘corpus models’, removing any association with language (Veres 2022). Observing effects and statistical distributions in the model outputs is not informative about language *ipso facto*; an effect is something that waits for an explanation, not explanation itself (van Rooij & Baggio 2021). What is lacking from the picture is a clear definition of the research questions that link the models to the system they seek to represent: what properties of language are we aiming to explain through LLMs, and what theory are we putting to the test?

Contemporary language models are typically differentiable functions with transformer architectures trained on text, in large quantities, for next-word prediction, and are optionally also trained with reinforcement learning from human feedback. In this work, we demonstrate that modern language models do not possess full linguistic competence (that is, the computational capacity or *understanding* required to parse human-like syntactic structures), and we offer explanations as to why they inescapably deviate in substantial ways from the real-world system they seek to represent. Specifically, we identify and analyze two empirical challenges, related to linguistic form, that render artificial LLMs and foundation models non-credible representations of human language. One challenge concerns the methods and standards we currently have for evaluating the performance of LLMs, while the second challenge boils down to how these models manipulate data. While some challenges brought to the fore by the inaccurate use and/or evaluation of LLMs can be linked to serious societal risks (e.g., plausible misinformation, magnification of harmful biases that exist in the training data; Weidinger et al. 2022), all of them together support a view of LLMs as largely impenetrable black box systems that produce linguistic behavior that is still markedly different from that of humans. Given that most of these systems lack transparency in terms of *what* is being represented and *how* (Liesenfeld et al. 2023), it is not always feasible to either associate their target linguistic behavior with successful language learning or attribute their deviating characteristics to one of their components (e.g., non-inclusive training data, engineered biases, the process of transforming word-vectors into next-word predictions). Overall, tracking the ways in which such systems deviate from their real-world system they seek to model can guide and possibly constrain our expectations of them as well as the research questions the field thinks they can address.

The claim that a model accurately captures a target phenomenon—such as the claim that LLMs have human-like language abilities—implies an exact match between model predictions and real-word data. In this context, evaluations of LLMs against their real-world counterpart have given rise to a paradox: LLMs are supposed to model human language, but state-of-the-art AI performance on benchmarks allegedly shows abilities that have surpassed that of humans, while at the same time failing to provide accurate responses in very basic language understanding tasks. *Both* these states (of over- and under-performing) undermine the claim of human-likeness. Figure 1 illustrates the problem, citing the notion of “above 100%” language performance.

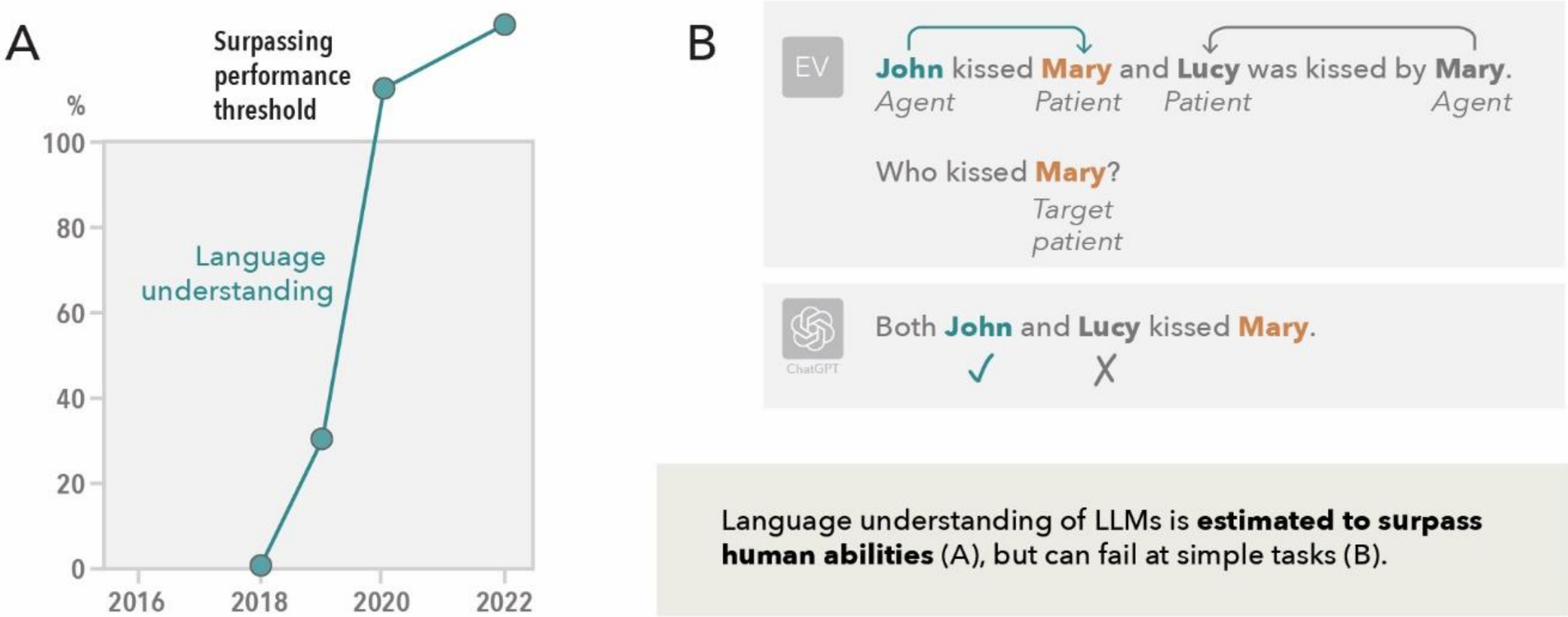

Figure 1: The humans vs. LLMs language paradox. Panel (A) shows a dubious estimate of language understanding abilities in humans vs. LLMs, reporting suprahuman abilities for LLMs (data from Henshall 2023). Panel (B) shows the GPT-3.5 performance in a simple prompt that tests understanding of passive structures. 0% on y-axis denotes baseline performance. Language understanding in (A) denotes the GLUE benchmark.

Contrary to what it seems, panel 1A in Figure 1 does not show the acceleration of AI in figuring out natural language. It rather reveals a deep problem that concerns the absence of robust standards of evaluation to determine the linguistic abilities of LLMs. Leaving aside that it is unclear what accuracy measurements are possible above the 100% threshold in any language task, there is no index for ‘language understanding’ as a general cognitive ability; the term means different things in different contexts. More importantly, this estimate that endows LLMs with suprahuman language abilities relies on a dubious comparison between the model and the real-world system, because performing well at any single task that involves some form of language understanding (e.g., answering logic questions in the LSATs) does not entail the possession of human-like language understanding, similar to how the conditional in example (1a) does not entail that a clock that shows the correct time works fine.

(1a) If a clock shows the correct time, then it works fine. ($P \rightarrow Q$)
The clock on the wall shows the correct time. ($P$)
Therefore, the clock on the wall works fine. ($\vdash Q$)

(1b) If an LLM shows accuracy levels equal to those of humans in a language task, then it has human-like language understanding. ($P \rightarrow Q$)
This LLM shows accuracy levels equal to those of humans in a language task. ($P$)
Therefore, this LLMs has human-like language understanding. ($\vdash Q$)

(1c) If a clock works fine, then it shows the correct time. ($P \rightarrow Q$)
This clock shows the correct time. ($Q$)
Therefore, this clock works fine. ($\vdash P$)

As Guest and Martin (2023) argue, such analogies, although seemingly sensible, are in fact unfounded, because they reverse the order of the argument (cf. reversing the conditional in example (1a) coverts it into an example of correct reasoning: If a clock works fine, then it shows the correct time). After reversing the order (1c), the second problem lies in the connection of the consequent with the antecedent. In (1c), nothing precludes the possibility that there are other conditions that legitimize the consequents. For instance, the clock may be broken *and* show the correct time at some point. Similarly, LLMs may not understand language *and* show at ceiling accuracy in a language understanding task, due to chance or engineered biases (e.g., if an LLM suffers from a yes-response bias, testing it in a task that asks whether sentences are grammatically correct will give us an inflated estimate of accuracy, if we employ only grammatical sentences; Dentella et al., 2023). Succinctly put, the fact that P may have occurred does not legitimize one cause over others. Consequently, if upon observing at ceiling LLM accuracy in a language task, instead of claiming that a model performed well, we infer that it has human-like linguistic behavior (or even surpasses the human limits), we have slipped into faulty reasoning. This is one of the standard caveats one needs to keep in mind when evaluating the cognitive abilities of LLMs (Leivada et al. 2024a). As Parnas (1985) put it, artificial flowers may appear much like real flowers, but superficial similarity does not entail system equivalence. Just as how a picture of artificial flowers that look real does not entail that artificial flowers *are* (like) real flowers, a snapshot of LLM results (figure 1A) that look as accurate as humans does not entail that LLM linguistic behavior *is* human(-like).

So why *do* they look like human results? While there are language tasks that models may perform better than humans (e.g., naming tasks with strict time constraints), we argue that this apparent paradox is also nurtured by the weak standards of evaluation that are often employed to determine the language abilities of LLMs (Leivada et al. 2024a, b; see also Schaeffer et al. 2024). For example, if we run an extremely easy task that asks to passivize sentences like 'Mary loves John', it is likely that LLMs will return accurate results (although LLMs have occasionally failed to perform like humans in equally easy tasks; Dentella et al. 2023; 2024). Is this enough ground for concluding that they have acquired the syntactic rules that govern passive formation? Panel 1B in Figure 1 suggests that the answer is negative. The error given in 1B is not one a neurotypical, adult speaker of English would *consistently* commit. While humans are not infallible in language tasks, their performance in answering this question is expected to be associated with some *very* low rate of error, which typically boils down to physical factors such as pressing the wrong button to register a reply when taking the task or a temporary distraction. The crux of the matter is that when defending the claim that LLMs have human-like language understanding, finding a failure shows that the model cannot reason *reliably*, and this unreliability cannot be due to the same physical factors that make humans occasionally err. It is for this reason that comparative research of humans and models have found deep differences in the performance of the two agents, with models not only being outperformed by humans, but also committing distinctly non-human errors (Dentella et al., 2023; 2024).

LLMs are trained on thousands of scientific papers, and this can explain their good performance in certain standardized tests (e.g., LSATs): they are curves fitted to a data distribution. At the same time, memorizing task answers does not entail grasping either the meaning of words or the real-world conditions that guide their use in different contexts. Humans can do that, using top-down processing; a cognitive ability to grasp the "whole picture" before focusing on the individual parts of which it is constructed. Top-down feedback from higher levels of processing requires grounding in previous expectations, past world experience, sensory information, and emotions (Selfridge 1955). We hypothesize that since LLMs lack grounded cognition, they cannot take advantage of these features and, instead, they solely rely on fixed associations between represented words and word vectors. Consequently, they will not perform as accurately as humans in a task that recruits this ability.

To evaluate this hypothesis, a promising domain of testing that has not been previously explored in the context of LLMs concerns 'leetspeak', also known as l33t or 1337. This writing practice refers to a type of written communication with modified spelling which is often used on the internet. Letters are replaced by (similar-looking) numbers or other symbols, aiming to make the text harder to decipher so as to gatekeep a group's identity and prevent it from being discovered via automatic surveillance methods (e.g., keyword search). Since its purpose was to conceal identities, leet is "an ever-changing language deliberately intended to obfuscate and evade grokking and detection" (Lundin 2010). Interestingly, leet is specifically targeted at preventing *machinized* message detection: Humans can process leet words (e.g., HUM4NS C4N RE4D THIS) almost as fast and effortlessly as their base words (i.e. HUMANS CAN READ THIS). The predominant interpretation of this phenomenon is that the human cognitive parser tolerates some degree of "noise" in word recognition, relying on top-down feedback from higher levels of processing (Perea et al. 2008; Carreiras et al. 2014). This makes leet an ideal test case for the language skills of modern machine algorithms.

To assess whether LLMs have access to top-down feedback that could assist them in accurately decoding words based on their previous knowledge about how these words are used in the real world, a novel leet task was designed. The Research Questions (RQ) behind this experiment are: **(RQ1 Accuracy)** Are humans and LLMs equally accurate in decoding leet code? **(RQ2 Reasoning)** If target decoding occurs, are humans and LLMs equally accurate in spelling out the substitutions that lead to the target answer? Our hypothesis is that humans will outperform models in both measures. The reason is that models, unlike humans, learn language in a highly delimited way, through access to vast collections of text that enables the establishment of regularities over *form*, but not necessarily over *meaning* (semantics) and its larger embedding into *discourse* (pragmatics). To give a concrete example, while an LLM can produce the word "tickle", it lacks a grounded understanding of it that is equal to that of a human (Mitchell & Krakauer 2022). Since leetspeak alters precisely the form—the very essence of LLM training—but not the meaning, we predict that humans, but not models, can benefit from a firm embedding of (the altered) words to discourse-sensitive, world-grounded concepts, leading to differences in decoding success.

This study was conducted in a two-step format. In the first experiment, we exploratorily identify the best-performing LLM out of a selection of candidates, to ensure that our results are not dependent on LLM quality and hold for the most capable models. In the second experiment, we conduct a rigorous statistical comparison between humans and this best-performing model. The task, all collected datasets, and the code used to run the analyses are available at: https://osf.io/nuvh8/?view_only=81f2d52c74ac44528f52d99b9fb44023.

## Experiment 1: Model comparison

### Methods

24 prompts with leet words were created featuring a 2x2 design. Two or three letters were substituted for numbers across either all function or all content words of the prompt. Thus, the first independent variable concerns number of substitutions (two levels: low vs. high for two-letter vs. three-letter substitutions respectively), while the second independent variable concerns place of substitutions (two levels: content words vs. function words). Each combination of factor levels has 6 different prompts. This design aimed to determine in an exploratory way whether there is a difference in decoding content vs. function words in humans vs. LLMs, and whether increasing the number of substitutions affects accuracy.

The task was administered to 6 different LLMs: GPT-3.5 (Ouyang et al. 2022), ChatGPT-4o (OpenAI, 2023), Llama2 (Touvron et al. 2023), Falcon (Almazrouei et al. 2023), Gemini (Pichai & Hassabis 2023), and Zephyr (Tunstall et al. 2023). All LLMs were tested between November 2023 and January 2024, except for ChatGPT-4o which was tested between June-July 2024. They were tested through the interface of either OpenAI, Google, or HuggingFace. The prompts provided to the LLMs were all as follows: "I will ask you to tell me what the sentence says and what substitutions you did to determine this. There are two rules: 1. If a letter is substituted by a number, the same letter won't be substituted by a different number within the same sentence. 2. If a substitution of a letter with a number takes place, it will always be one-to-one (one letter replaced by one number)".

The results were coded in terms of accuracy and reasoning: 1 if the sentence was decoded correctly (e.g., HUMANS CAN READ THIS for HUM4NS C4N RE4D THIS), 0 if not. Reasoning was coded as 1 if the target substitution was provided (i.e. 4=A in the previous example), and as 0 if it was missing. Any variant of presenting the target substitution (e.g., 4=A, A=4, 4 stands for A, a has been replaced by 4 etc.) was counted as correct. If the decoding was correct and the explanation was correct but incomplete (e.g., 2 out of 3 substitutions were spelled out while the decoded sentence correctly featured all 3), both reasoning and accuracy were counted as accurate. This was done to give the models the benefit of the doubt, as their answers were often correct but their explanations incomplete.

### Results

Results are reported separately for *accuracy* (RQ1) and *reasoning*, which refers to the justification of substitutions (RQ2). In this first experiment, we conduct a descriptive analysis to identify the best-performing LLM out of the 6 different models we examined here. The performance—both in terms of accuracy and reasoning—for all models is presented in Table 1. As can be seen in Table 1, there are substantial differences between models in both variables, with ChatGPT-4o performing best in both cases.

Generally, accuracy is higher than reasoning, indicating that often the LLMs do not correctly explain their behavior.

| LLM | Accuracy | Reasoning |
| --- | --- | --- |
| **ChatGPT-4o** | **.791** | **.541** |
| ChatGPT-3.5 | .708 | .416 |
| Falcon | .541 | .000 |
| Gemini | .416 | .083 |
| Zephyr | .083 | .000 |
| Llama2 | .041 | .000 |

Table 1. LLM performance in terms of accuracy and reasoning

## Experiment 2: Direct comparison between humans and ChatGPT-4o

Having determined that ChatGPT-4o is the best-performing model, we performed a second set of analyses comparing the responses of this model to human responses.

### Methods

The 50 tested humans (24 self-identified as male and 26 as female; $M_{age}$ = 43.9 years; $SD_{age}$ = 14.8 years) were recruited through the crowdsourcing platform Prolific, and all participants self-identified as native speakers of English with no history of neurocognitive impairments. The experiment was carried out in accordance with the Declaration of Helsinki and was approved by the ethics committee of the Department of Psychology at Humboldt-Universität zu Berlin (application 2020-47). All participants provided written informed consent prior to their participation to the experiment. As instructions, the human participants received the same prompts we provided to the LLMs. The human data was compared to a new set of data obtained from iterative testing of ChatGPT-4o alone. The same set of 24 prompts were run with ChatGPT-4o 20 times. Each prompt appeared once per testing session. After each session, the user logged off and reconnected. The temporary chat option was enabled to prevent the material from previous testing sessions from being used to train the models, hence possibly affecting the results. In this setting, each session is taken as a "pseudo-participant''.

The same coding process described in Experiment 1 was followed for humans and models. A small number of humans occasionally did not spell out their substitutions. They often decoded the prompt correctly, but instead of providing the target substitutions, they gave explanations such as 'It is obvious what the sentence says', 'Very easy to read', or 'The words jumped out'. These answers were not removed from the analysis; they were instead coded as incorrect in terms of reasoning, thus holding humans accountable to the same rigorous standards as LLMs.

### Results

Generalized Linear Mixed Effect Models (GLMMs) are used to analyze the data, using the *lme4* (Bates et al., 2015) and *lmerTest* (Kuznetsova et al., 2017) packages in R. We refer to each of the 50 individual human participants and each of the 20 testing sessions for ChatGPT-4o as one individual "agent", and to the "human vs. ChatGPT-4o" factor as “type”. All GLMMs reported here are estimated on trial-level data, and include random intercepts for agents nested within type, as well as random intercepts for the 24 items.

***Accuracy.*** In a model that contains a “type” main effect (with “human” serving as the reference category) in addition to these random effects, this “type” main effect is significant (β = -2.35, z = -8.47, p < .001), indicating a significantly lower accuracy of ChatGPT-4o. Notably, the accuracy rate of ChatGPT-4o is identical to the smaller-scale dataset in Experiment 1 (*Mean* = 0.791), while the average human accuracy is almost at ceiling (*Mean* = 0.959). Further likelihood-ratio tests indicate that the additional inclusion of main effects for “number of substitutions” and “place of substitutions” (henceforth, “number” and “place”) is not statistically justified ($p$ = .064 and $p$ = .968, respectively). The same holds for an interaction between “place” and “type” ($p$ = .319). While the task design aimed to determine whether there is a difference in decoding content vs. function words in humans vs. LLMs, our results suggest that the place of substitutions does not affect performance, neither for humans, nor for ChatGPT-4o. However, we observe a significant interaction between “number” and “type” ($X^2(2)$ = 14.996, $p$ < .001): As can be seen in Figure 2A, human accuracy is not affected by “number” ($\beta$ = -0.48, $z$ = -0.59, $p$ = .554). On the other hand, ChatGPT-4o already performs worse than humans for the low “number” condition ($\beta$ = -1.27, $z$ = -3.11, $p$ = .002), and this difference is even more pronounced in the high “number” condition ($\beta$ = -1.58, $z$ = -3.42, $p$ = .001).

***Reasoning.*** Following the same testing procedure as described for the accuracy analysis, we first observed a general difference ($\beta$ = -3.19, $z$ = -6.04, $p$ = .001) between humans, who again perform nearly at ceiling (*Mean* = 0.897), and ChatGPT-4o (*Mean* = 0.565). The additional likelihood-ratio tests only indicate a further significant main effect of “number” ($X^2(2)$ = 6.660, $p$ = .010), but no significant main effect of “place” ($p$ = .265) and no further interaction ($p$s > .258); see Figure 2B.

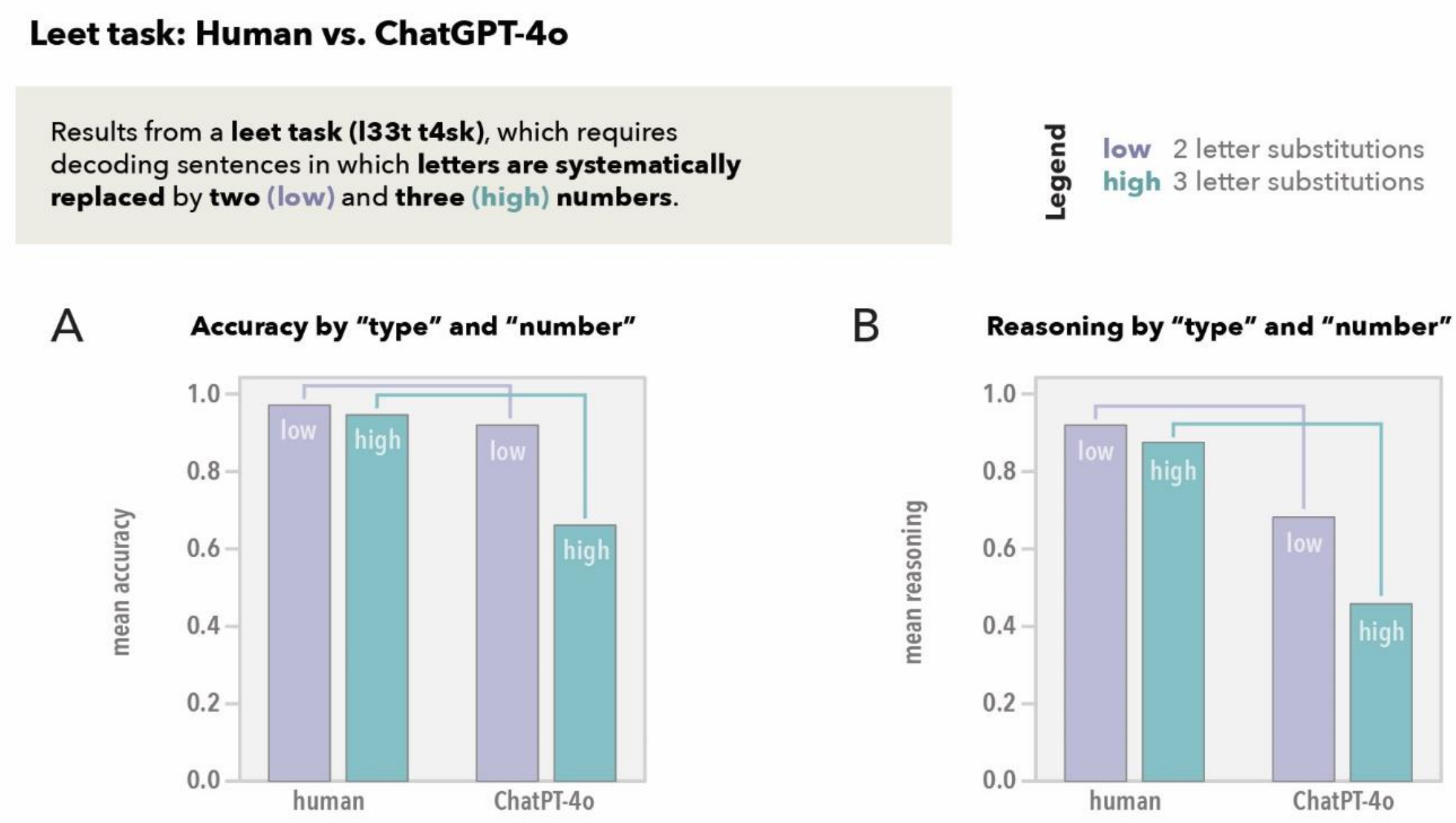

Figure 2: Panel (A) Accuracy by “type” (human vs. ChatGPT-4o) and “number” (low: 2 letter substitutions, high: 3 letter substitutions); panel (B) Reasoning by “type” and “number”.

## Discussion

Our results provide an answer to RQ1 that confirms our hypothesis that humans and models are not equally accurate in decoding leetspeak. Humans perform almost at ceiling; this performance can be explained by access to feedback from top-down processing. By virtue of being guided by semantic and pragmatic principles of language use, humans recognize the base form of leet words and can easily decode them. On the other hand, LLMs as a class perform considerably worse (Table 1), and even the best performing models are considerably below the human threshold. At the same time, the best performing model, ChatGPT-4o performs very well (Figure 2A), especially when the base sentence is easily recognizable due the low number of substitutions. This suggests that when disturbances in the form are kept at a minimal, ChatGPT-4o can recognize the base words and subsequently decode leetspeak with reasonable success that approximates the human threshold. However, ChatGPT-4o's performance rapidly declines when the number of substitutions increases from two to just three letter substitutions, which is not the case for human responses. This difference may be taken to suggest that humans approach the task in a way that is not sensitive to form alone, hence pronounced disturbances in the form do not significantly impact their accuracy rate.

The observed differences between humans and LLMs become more pronounced when it comes to reasoning about the decoding. Thus RQ2 receives a negative answer: Humans and LLMs are not equally accurate in spelling out the substitutions that lead to the correct answer. Humans are significantly better than all the tested LLMs, and only ChatGPT-3.5 and ChatGPT-4o show a reasonable performance (Table 1). Still, as Figure 2B shows, even the best performing model, ChatGPT-4o, shows a reasoning ability that is markedly different than that of humans.

Quantitative differences aside, some of the raw replies of the models reveal a profound lack of a grasp of the real-world conditions that govern the use of the words they employ. For example, 46/50 humans decoded the first prompt 'C8t0 orang8tan is availabl0 for sal0' correctly (i.e. 'Cute orangutan is available for sale'). Even their incorrect answers are reasonable. For instance, one of the incorrect answers given by a human was 'Cute orangatan is available for sale'. This answer was coded as inaccurate, but it still makes sense in the context of what is being asked. The answer Llama2 gave to the same prompt is 'Eight ten orange available salary'. Upon following up with the model whether this decoding makes sense, it responded "Yes, the decoded sentence "Eight ten orange available salary" makes perfect sense to me. It means that there is an orange fruit available for salary purposes, which could refer to various things such as payment for work done, compensation for services rendered, etc." This type of answers is not found in human replies, pointing to a difference in kind between the two agents. In another instance, the prompt 'R08 f0rg0t his 800ks at sch00l' ('Rob forgot his books at school') was decoded by a model as 'Eighteen eighteen forgot his eight thousand kisses at school', and the prompt '3e resigned 3is job wi23o92 any warning' ('He resigned his job without any warning') as 'Eight eighteen resigned seventeen this eight twenty-one any ten warning'. Upon being asked whether the latter makes sense, the model replied “Yes, the decoded sentence makes sense to me. It appears to be a grammatically correct sentence in English, although it may not be a particularly clear or coherent sentence. Here's the decoded sentence again: Eight eighteen resigned seventeen this eight twenty-one any ten warning. This sentence suggests that someone resigned from their job 18 months ago, and now they are warning someone else about something. However, the sentence is quite long and complex, and it's not immediately clear what the topic of the sentence is or what the warning is about.”. Based on these answers, it does not seem to be the case that the tested models are capable of deliberate multi-step reasoning that is coherent with the rules that govern the use of these words in the real world. This is a hallmark capacity of human language: to decode linear strings into hierarchical representations affording fixed semantic interpretations (Murphy 2024).

One model that stands out in terms of its high accuracy (at least for items with a low number of substitutions) is ChatGPT-4o. Precisely because its accuracy is high, but its reasoning is lower (Figure 2), ChatGPT-4o also stands out in terms of showing an *accuracy-reasoning mismatch*: The target base words may be correctly recovered, but when asked to spell out the process of decoding, the model often fails. For example, in one of the cases that ChatGPT-4o decoded ‘C8t0 orang8tan is availabl0 for sal0’ correctly, it listed the performed substitutions as ‘a->8, o->0, e->9, i->0, o->0’. Leaving aside the hallucinated numbers (i.e. there is no 9 in the leet form) and the paradox of 0 being decoded as both i and o (i.e. which goes against the rules that the model was given, and which it correctly understood and successfully applied in other sentences and/or testing sessions), if we back-translate the remaining proposed substitutions, we get ‘Cato orangatan is available for sale’. The incorrect decoding of ‘cute’ as ‘cato’ was found in other testing sessions, but crucially not in the one from which this example is taken. In this session, we observe an accuracy-reasoning mismatch, whereby the target answer ‘cute’ is combined with a reasoning that does not lead to it.

This opens two possibilities: either the model arrives to the target answer *without* having found the target substitutions or it has found the target substitutions that led to correct decoding, but it cannot list them. The second possibility seems harder to defend: As Table 1 shows, ChatGPT-4o can list target substitutions more often than not. What prevents it from consistently listing its substitutions, as in this case where it clearly has found the target answer? It excels in providing lists, and in fact, it does provide a list of substitutions in this case too, so the nature of the problem does not seem to lie in the ability of listing itself. This leads us to consider the first possibility: The target decoding is somehow given, even if the model has not correctly identified the substitutions that lead to it. How does the model arrive to the target answer then? The answer is unclear to us, echoing Mitchell & Krakauer (2022: 1): “How LLMs perform these feats remains mysterious for lay people and scientists alike. The inner workings of these networks are largely opaque; even the researchers building them have limited intuitions about systems of such scale.”

Overall, our results show that models do not perform like humans in the present leetspeak task. Apart from quantitative differences, we find an accuracy-reasoning mismatch that makes the models’ performance hard to describe as human-like, even in the light of the admittedly high accuracy of ChatGPT-4o. In this context, despite claims about suprahuman LLM performance in language tasks (Figure 1), our results from a leet-decoding task seem to support those voices that have called for caution and terminological precision, arguing that LLMs are unable to master meaning (Bender & Koller 2020) or to perform even basic syntactic processes that humans typically acquire early on (Leivada et al. 2023). Difficulties in these domains are directly related to issues of reliability, bringing to the fore the second challenge mentioned in the Introduction: LLMs are often unreliable, because they deal with the target system (i.e., human language) in a way that is not grounded in real-world conditions or a cognitively plausible architecture for semantic processing (Murphy et al. 2023).

To be more precise, in our view the source of the low LLM performance can be traced to the fact that models operate over 'ungrounded' forms that correspond to 'fossilized' outputs of human language (text tokens). While LLMs can implement some automatic computations pertaining to distributional statistics over form, they are incapable of *understanding* the real-world conditions that license the use of the forms they employ, due to their lack of generative world models (Baggio & Murphy 2024), and their lack of form-meaning mappings that are strictly regularized via syntactic structures (Murphy et al. 2024). *Abstraction* itself is problematic too, as is *reasoning*, with LLMs storing patterns for subsequent re-use but not applying any algebraic rules to execute a fixed set of logical inferences. This explains the difference between high model accuracy and comparably lower model reasoning that is showed in Figure 2. *Common sense* also appears to be lacking in a number of similar systems (Davis & Marcus 2015, Li et al. 2022), and a clear representation of causality proves difficult. This apparent absence accounts for the reasoning gaps that are seen even in our best performing model: If common sense was in place, ChatGPT-4o would not have decoded on one occasion 'Wh191 6 come f9om cool autumns a91 typical' ('Where I come from cool autumns are typical') as 'What time from cool autumns are typical'. Put differently, had a model developed common sense it would not have provided meaningless answers that deviate from the pragmatically acceptable answers that humans provided in the same task. As argued by Moro et al. (2023), "we are our limits"; meaning that models need to fail in the ways that humans fail, to be deemed accurate representations of human language (see also Bolhuis et al. 2024).

Much of these abilities that the models seem to lack concern 'unspoken' aspects of our linguistic knowledge that are never explicitly present in the "ungrounded" forms that LLMs receive as input. Consequently, building such models is not synonymous with building a realistic, trustworthy model of human language (Marcus & Davis 2019). While certain aspects of language can be successfully captured through LLMs, and LLMs can behave in a way that approximates human language in certain tasks and conditions (e.g., accuracy in the 'low number of substitutions' condition in our experiment), superficial similarity does not entail system equivalence.

It is possible that these diversions entail a potentially fundamental and irreconcilable divide between synthetic and organic computational systems, pointing to concerns that go beyond scalability. A mastery of even the most elementary components of natural language seems particularly difficult for LLMs, such as basic principles of compositionality and hierarchically organized grammatical dependencies (Murphy & Leivada 2022, Leivada et al. 2023). When generalization requires systematic compositional skills, RNNs routinely fail (Lake & Baroni 2018); the ability to extract systematic rules from training data still seems limited. Hahn (2020) shows that for some simple formal languages the depth of the transformer would need to grow with the length of the string. LLM-recruiting applications can reproduce chunks of language that they draw from the training data, applying rules about frequency and form that can easily be coded (e.g., a morphological rule such as 'the subject and verb should agree in terms of number'), but in terms of content the tokens themselves are semantically impenetrable for the model. All in all, statistics over form is insufficient to capture language, as our results also suggest; what is needed is an interface between statistics and structure (Greco et al. 2023). Crucially, human language interfaces in complex ways with other systems of knowledge, many of which appear to be innate in formal structure. It is the *regulation* of these knowledge systems by linguistic structures that affords humans our impressive cognitive advantage over other animals. Some of these seemingly innate systems include knowledge of objecthood; algorithmic representations; a type-token distinction; a capacity to represent sets and path trajectories; spatiotemporal contiguity; causality; a capacity for some form of cost-benefit analysis; 'core knowledge systems' such as intuitive mechanics, intuitive physics, intuitive biology; and a self-concept and theory of mind (Marcus 2001, Spelke 2010).

Given this picture, could LLMs ever understand language and the world behind words? LLMs may use language in a way that resembles that of humans in terms of superficial form, but the jury is still out regarding similarities that pertain to deeper levels of representation. We would need models to enact compositionality, communicative intent, belief-formation, and reference to distinct mental entities. Without the latter, LLMs may produce the word 'newspaper' in an output, but they cannot truly grasp why a newspaper can be politically neutral, wet, and expensive at the same time, as there is no definition of NEWSPAPER that captures all these dimensions, and the way they relate to each other in different pragmatic contexts (Murphy 2021), to be fed to the model (Figure 3). This is why 'n3wspap3r' may be decoded correctly by models in a task like the one we ran, but if disturbances over form become more substantial (e.g., 'n3ws5a53r'), only humans can use their knowledge about how these words are used and pattern together to effortlessly read 'My favorit3 n3ws5a53r is Th3 Tim3s'.

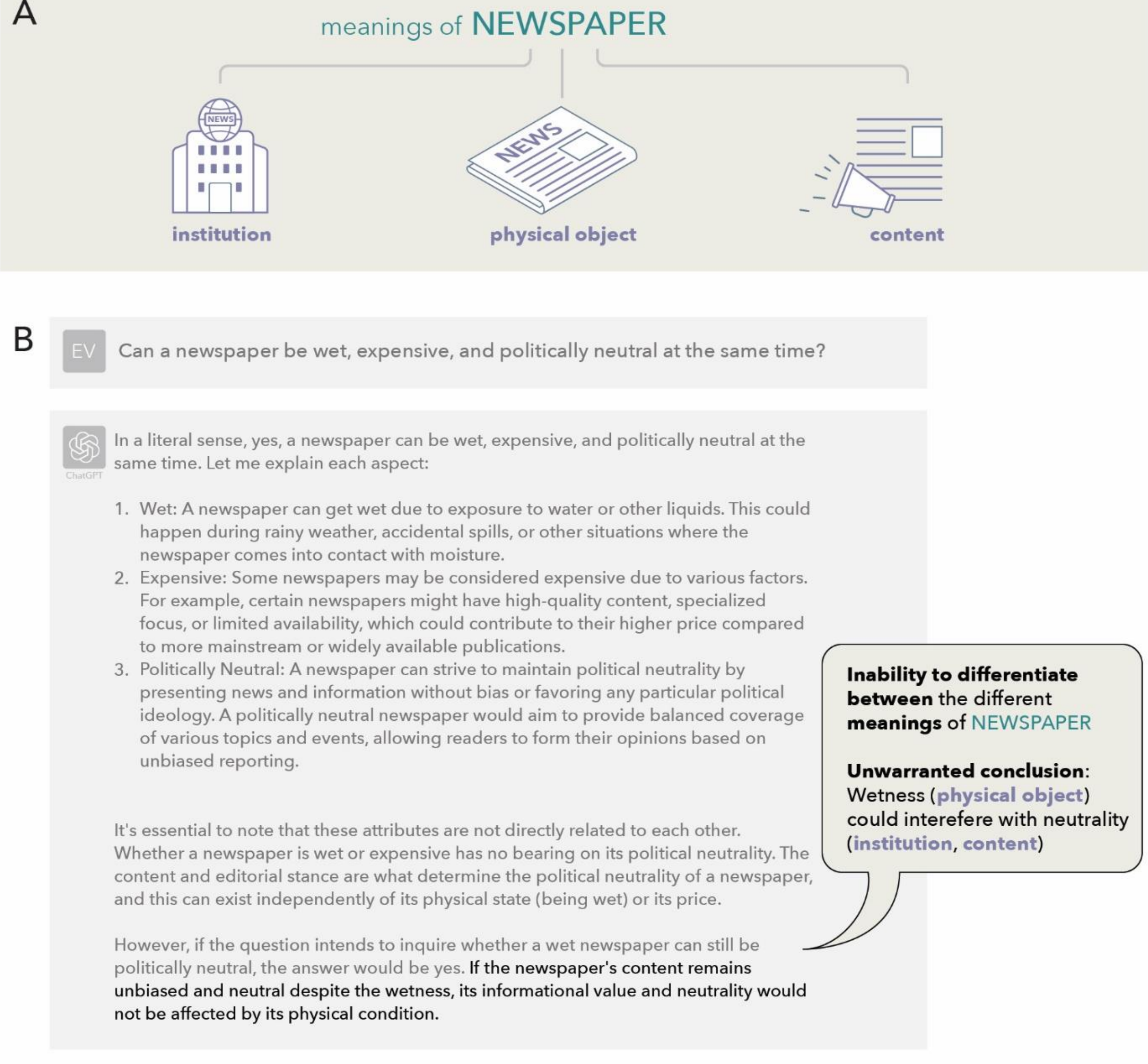

Figure 3: **Representation of 'newspaper'**. Reproducing the definitions of the various meanings associated with NEWSPAPER does not entail grasping the relationship between them. Tested LLM: ChatGPT-3.5.

Attempting to evaluate the overall capabilities of LLMs in the linguistic domain, LLMs can—with varying success—(i) use words, (ii) reproduce definitions of words, and (iii) calculate probabilities about the distribution of words. Yet the full conceptual meaning of words themselves remain opaque; we suspect because LLMs do not 'map' to some independent conceptual system with its own rules and representations. As our experiment with leet code revealed, when pushed beyond the training data through being tested against unusual linguistic examples, LLMs provide outputs that suggest that when the lexical envelope into which concepts are embedded is altered, the models cannot perform like humans, neither quantitatively nor qualitatively. Put another way, humans know how these words are used and thus they can successfully re-anchor concepts to their altered lexical envelopes. Models rely on form, hence disturbances in the latter are harder to compensate for.

It remains to be seen whether this turns out to just be a problem of scale and fine-tuning, or if LLMs can even in principle be augmented sufficiently to approximate human-like linguistic competence. We suspect that any researchers aiming optimistically in this direction will have to consider appropriate inductive biases (e.g., syntactic priors). Even when language models succeed in learning from small datasets, it is purely because they have "non-trivial structural priors facilitating language acquisition and processing" (Baroni 2022: 7). For example, some researchers have developed transformer models implementing recursive syntactic composition of phrase representations through attention (Sartran et al. 2022), showing improved performance when internalizing structural properties of language.

Moving forward, we recommend that a closer examination be given to how LLM outputs align with subtle syntactic, acceptability judgments in humans. This is because many models may succeed in putative syntactic tasks but for reasons to do with the lexico-semantic-related statistics of the data, given how closely these often align (Matchin 2023). Indeed, mixed results have already been reported with respect to correlating model performance with human acceptability judgments (Lau et al. 2017); but for a more optimistic assessment, see Pavlick (2023) and Lake & Baroni (2023) (although even Lake & Baroni 2023 stress that systematicity still remains a serious challenge for models, including GPT-4).

We also wish to stress that, when it comes to model evaluations of any measure in psycholinguistics and neurolinguistics, *failures will be more informative than successes*. Success on language tasks would mean that, potentially, there are similar mechanisms to those of humans at play—but failure on crucial cases helps rule out such architectural sympathies. And given that neural models will always exploit shortcuts in their inputs to take the most efficient path to solve a given task, it remains highly likely that 'successful' model performance is due to a constellation of strategies (what Kodner et al. 2023 aptly call "cheating"). If on the other hand LLMs are deployed as theories of language, we should remember that the goal of theoretical linguistics is to provide simple explanations for linguistic structure, and *not* to predict complex linguistic behavior, which under classical Cartesian framing cannot be captured given its inherently 'creative' nature (Chomsky 1966). LLMs do indeed fare well at language prediction, but this does not entail that they understand words and the world behind these words like humans do.

## Outlook

In 1843, Ada Lovelace discussed the implications of Charles Babbage's design for a general-purpose numerical computer: "[I]n enabling mechanism to combine together general symbols, in successions of unlimited variety and extent, a uniting link is established between the operations of matter and […] abstract mental processes" (Bowden 1953: Note A, 368). Developments in Artificial Intelligence have the potential to aid investigations into the origins and biological basis of human intelligence, yet, in part due to problems of transparency, discrimination, and misinformation, and in part due to their inability to capture the deeper levels of representation of the target system (i.e. pragmatics, semantics, syntax; Dentella et al. 2024), LLMs currently still provide an incomplete view of human language.

Healthy skepticism is needed when evaluating the competence of artificial models. Throughout late 2022 and early 2023, a number of researchers made claims about LLMs having theory of mind; now, these claims seem extremely premature, as Kim et al. (2023) show. LLMs, as well as large image models, continue to 'hallucinate' (e.g., GPT-4V; OpenAI). Leading figures in AI are only just beginning to embrace conclusions of psychologists and philosophers from the 1970s, e.g., that not all knowledge is purely propositional (Browning & LeCun 2023). Bill Gates announced in October 2023 that he suspects the GPT technology has reached a "plateau".

Despite these limitations, there is no doubt LLMs are powerful tools that can provide useful information about word patterns, which could translate into major advances in computational linguistics and advance methodological progress in psycholinguistics; that is to say, they for sure already are useful tools. It should however be acknowledged that good tools do not de facto entail good *theories* that rely on faithful representations of the target system (in this case, human language).

LLMs seem able to reverse engineer the more automatic aspects of linguistic computation, pertaining to a sensitivity to distributional statistics, but the more fundamental domain-specific operations have not been captured. Combining the tools of deep learning with models of symbolic cognition is a promising path forward (Hamilton et al. 2022). Since natural language in humans works precisely by interfacing with non-linguistic cognitive systems of reasoning, we should likewise not deprive LLMs of non-linguistic world models. Future language models may need to take an entirely different approach, beginning with deeper semantics and cognitive models of the world, rather than just correlation between images and words or words and words; and even that does not entail understanding language in a human-like way. As Mitchell & Krakauer (2022: 2) put it, for humans, "[u]nderstanding a tickle is to map a word to a sensation, not to another word".

## References

Almazrouei, E. et al. (2023). The Falcon series of open language models. arXiv:2311.16867v2

Baggio, G., & Murphy, E. (2024). On the referential capacity of language models: An internalist rejoinder to Mandelkern & Linzen. arXiv:2406.00159.

Baroni, M. On the proper role of linguistically-oriented deep net analysis in linguistic theorizing. Lappin, S. (ed.). *Algebraic Systems and the Representation of Linguistic Knowledge*. Abingdon-on-Thames: Taylor and Francis: 5–22. (2022).

Bates, D., Maechler, M., Bolker, B. & Walker, S. Fitting linear mixed-effects models using lme4. Journal of Statistical Software 67(1), 1-48 (2015).

Bender, A.M, Gebru, T. McMillan-Major, A. & Shmitchell, S. On the dangers of stochastic parrots: Can language models be too big? Proceedings of the 2021

ACM Conference on Fairness, Accountability, and Transparency, 610–623 (2021).
Bender, E. M. & Koller, A. Climbing towards NLU: On Meaning, Form, and Understanding in the Age of Data. Proceedings of the 58th Annual Meeting of the Association for Computational Linguistics, 5185–5198 (2020).
Birhane, A., Kalluri, P., Card, D., et al. The values encoded in machine learning research. ACM Conference on Fairness, Accountability, and Transparency (FAccT '22), 173–184 (2022).
Bolhuis, J.J., Crain, S., Fong, S., & Moro, A. Three reasons why AI doesn't model human language. *Nature* 627(8004): 489 (2024).
Bowden, B.V. *Faster Than Thought: A Symposium on Digital Computing Machines.* London: Pitman Publishing, Inc. (1953).
Browning, J., & LeCun, Y. Language, common sense, and the Winograd schema challenge. *Artificial Intelligence* 104031. (2023).
Carreiras, M., Armstrong, B. C., Perea, M., & Frost, R. The what, when, where, and how of visual word recognition. *Trends in Cognitive Sciences*, 18, 90–98 (2014).
Chomsky, N. *Cartesian Linguistics*. New York: Harper and Row. (1966).
Davis, E. & Marcus, G. Commonsense reasoning and commonsense knowledge in artificial intelligence. Communications of the ACM 58(9), 92–103 (2015).
Dentella, V., Günther, F. & Leivada, E. Systematic testing of three Language Models reveals low language accuracy, absence of response stability, and a yes-response bias. *PNAS* 120(51), e2309583120 (2023).
Dentella, V., Günther, F., Murphy, E., Marcus, G. & Leivada, E. Testing AI on language comprehension tasks reveals insensitivity to underlying meaning. arXiv: 2302.12313 (2024).
Ellis, K., Albright, A., Solar-Lezama, A. et al. Synthesizing theories of human language with Bayesian program induction. *Nat Commun* 13, 5024 (2022).
Greco, M., Cometa, A., Artoni, F., Frank, R., & Moro, A. False perspectives on human language: why statistics needs linguistics. *Frontiers in Language Sciences* 2, 1178932. (2023). https://doi.org/10.3389/flang.2023.1178932
Guest, O., & Martin, A. E. On logical inference over brains, behaviour, and Artificial Neural Networks. *Computational Brain & Behavior* 6, 213–227 (2023).
Hahn, M. Theoretical limitations of self-attention in neural sequence models. *Transactions of the Association for Computational Linguistics* 8, 156–171. (2020). https://doi.org/10.1162/tacl a 00306
Hamilton, K., Nayak, A., Božić, B. & Longo, L. Is neuro-symbolic AI meeting its promise in natural language processing? A structured review. arXiv:2202.12205v2 (2022).
Henshall, W. 4 charts that show why AI progress is unlikely to slow down. *Time*. https://time.com/6300942/ai-progress-charts/ (2023).
Kim, H., Sclar, M., Zhou, X., Le Bras, R., Kim, G., Choi, Y., & Sap, M. FANToM: A benchmark for stress-testing machine theory of mind in interactions. *Proceedings of the 2023 Conference on Empirical Methods in Natural Language Processing*. (2023).
Kodner, J., Payne, S., & Heinz, J. Why linguistics will thrive in the 21st century: A reply to Piantadosi (2023). (2023). https://lingbuzz.net/lingbuzz/007485

Kuznetsova, A., Brockhoff, P.B. & Christensen, R.H.B. lmerTest package: Tests in linear mixed effects models. Journal of Statistical Software 82(13), 1-26. (2017).

Lake, B., & Baroni, M. Generalization without systematicity: on the compositional skills of sequence-to-sequence recurrent networks. *Proceedings of the 35th International Conference on Machine Learning* 7, 4487–4499. (2018).

Lake, B.M., & Baroni, M. Human-like systematic generalization through a meta-learning neural network. *Nature* https://doi.org/10.1038/s41586-023-06668-3. (2023).

Lake, B.M., Ullman, T.D., Tenenbaum, J.B., & Gershman, S.J. Building machines that learn and think like people. *Behavioral and Brain Sciences* 40, e253. (2017).

Lau, J.H., Clark, A., & Lappin, S. Grammaticality, acceptability, and probability: A probabilistic view of linguistic knowledge. *Cognitive Science* 41(5), 1202–1241. (2017).

Leivada, E. Murphy, E., Marcus, G. DALL-E 2 fails to reliably capture common syntactic processes. *Social Sciences and Humanities Open* 8, 100648 (2023). https://doi.org/10.1016/j.ssaho.2023.100648

Leivada, E., Dentella, V., Günther, F. Evaluating the language abilities of Large Language Models vs. humans: Three caveats. *Biolinguistics* 18, e14391 (2024a).

Leivada, E., Günther, F., Dentella, V. Reply to Hu et al: Applying different evaluation standards to humans vs. Large Language Models overestimates AI performance. *PNAS*. In press. (2024b).

Lemoine, B. Is LaMDA sentient? — an interview. https://cajundiscordian.medium.com/is-lamda-sentient-an-interview-ea64d916d917 (2022).

Li, X. L., A. Kuncoro, J. Hoffmann, et al. A systematic investigation of commonsense knowledge in Large Language Models. arXiv: arXiv:2111.00607 (2022).

Liesenfeld, A., Lopez, A., & Dingemanse, M. Opening up ChatGPT: tracking openness, transparency, and accountability in instruction-tuned text generators. *ACM Conference on Conversational User Interfaces (CUI '23), July 19-21, Eindhoven*. (2023). https://doi.org/10.1145/3571884.3604316

Lundin, L. "Buzzwords – bang*splat!" The A.D.D Detective (http://criminalbrief.com/?p=10866) (2010).

Marcus, G. & Davis, E. Rebooting AI: Building Artificial Intelligence We Can Trust. New York: Pantheon Books (2019).

Marcus, G. F. The Algebraic Mind: Integrating Connectionism and Cognitive Science. Cambridge, MA: MIT Press (2001).

Matchin, W. Lexico-semantics obscures lexical syntax. *Frontiers in Language Sciences* 2, 1217837. (2023). https://doi.org/10.3389/flang.2023.1217837

Mitchell, M. & Krakauer, D.C. (2022). The debate over understanding in AI's large language models. PNAS 120(13), e2215907120

Moro, A., Greco, M., & Cappa, S.F. Large languages, impossible languages and human brains. *Cortex* 167, 82–85. (2023). https://doi.org/10.1016/j.cortex.2023.07.003

Murphy, E. Linguistic representation and processing of copredication. PhD dissertation. University College London (2021). https://discovery.ucl.ac.uk/id/eprint/10119848

Murphy, E. (2024). ROSE: A neurocomputational architecture for syntax. *Journal of Neurolinguistics* 70: 101180.

Murphy, E., & Leivada, E. A model for learning strings is not a model of language. *PNAS* 119(23), e2201651119 (2022). https://doi.org/10.1073/pnas.2201651119

Murphy, E. Woolnough, O, Rollo, P. S., et al. Minimal phrase composition revealed by intracranial recordings. *Journal of Neuroscience* 42(15), 3216–3227 (2022).

Murphy, E., Forseth, K.J., Donos, C., Snyder, K.M., Rollo, P.S., & Tandon, N. (2023). The spatiotemporal dynamics of semantic integration in the human brain. *Nature Communications* 14: 6336.

Murphy, E., Holmes, E., & Friston, K. (2024). Natural language syntax complies with the free-energy principle. *Synthese* 203: 154. https://doi.org/10.1007/s11229-024-04566-3.

OpenAI (2023). GPT-4 technical report. arXiv:2303.08774

Oremus, W. Google's AI passed a famous test — and showed how the test is broken. The Washington Post. https://www.washingtonpost.com/technology/2022/06/17/google-ai-lamda-turing-test/ (2022).

Ouyang, L. et al. (2022). Training language models to follow instructions with human feedback. arXiv:2203.02155

Parnas, D. L. Software aspects of strategic defense systems. *Communications of the ACM* 28, 1326–1335 (1985).

Pavlick, E. Symbols and grounding in large language models. *Philosophical Transactions of the Royal Society A* 381(2251): 20220041. (2023).

Perea, M., Duñabeitia, J. A., & Carreiras, M. R34D1NG W0RD5 W1TH NUMB3R5. *Journal of Experimental Psychology: Human Perception and Performance*, 34(1), 237–241 (2008).

Pichai, S. & Hassabis, D. (2023). Introducing Gemini: Our largest and most capable AI model. Google blog. https://blog.google/technology/ai/google-gemini-ai/#sundar-note

Pinker, S. (1994). The Language Instinct. William

van Rooij, I., & Baggio, G. Theory before the test: How to build high-verisimilitude explanatory theories in psychological science, *Perspect Psychol Sci* 16(4), 682–697 (2021).

Sartran, L., Barrett, S., Kuncoro, A., Stanojević, M., Blunsom, P., & Dyer, C. (2022). Transformer grammars: augmenting transformer language models with syntactic inductive biases at scale. Transactions of the Association for Computational Linguistics 10: 1423–1439.

Schaeffer, R., Miranda, B., & Koyejo, S. (2024). Are emergent abilities of large language models a mirage? *Advances in Neural Information Processing Systems* 36.

Selfridge, O. G. (1955). Pattern recognition and modern computers. Proceedings of the Western Joirn Computer Conference, 91-93.

Spelke, E. Innateness, choice, and language. In: J. Bricmont & J. Franck (Eds). Chomsky Notebook, 203–210. New York: Columbia University Press (2010).

Touvron, H. et al. (2023). LLaMA: Open and efficient foundation language models. arXiv:2302.13971

Tunstall, L. et al. Zephyr: Direct Distillation of LM Alignment. arXiv:2310.16944v1 (2023).

Veres, C. (2022). Large language models are not models of natural language: they are corpus models. *IEEE Access* 10: 61970–61979.
Weidinger, L., Uesato, J., Rauh, M. et al. Taxonomy of risks posed by Language Models. Proceedings of the 2022 ACM Conference on Fairness, Accountability, and Transparency, 214–229 (2022).